\documentclass[11pt,a4paper]{article}
\usepackage[hyperref]{naaclhlt2019}
\usepackage{times}
\usepackage{latexsym}
\usepackage{amsmath}
\usepackage{bbm}
\usepackage{amssymb}
\usepackage{url}
\usepackage{tabularx}
\usepackage{cleveref}
\crefformat{section}{\S#2#1#3} 
\crefformat{subsection}{\S#2#1#3}
\crefformat{subsubsection}{\S#2#1#3}
\usepackage{graphicx}
\usepackage{tikz}
\usetikzlibrary{patterns}
\usepackage{subfig}
\usepackage{comment}

\usepackage{tabularx}
\newcommand\numberthis{\addtocounter{equation}{1}\tag{\theequation}}
\usepackage[noend]{algpseudocode}
\usepackage{algorithm}
\algrenewcommand\algorithmicindent{1em}
\usepackage[font=footnotesize,labelfont=bf]{caption}
\makeatletter
\algrenewcommand\ALG@beginalgorithmic{\small}
\makeatother
\usepackage{booktabs}
\usepackage{tikz-qtree}
\newcommand{\boldx}{\mathbf{x}}
\newcommand{\boldw}{\mathbf{w}}
\newcommand{\boldW}{\mathbf{W}}

\newcommand{\boldg}{\mathbf{g}}

\newcommand{\boldz}{\mathbf{z}}
\newcommand{\bolde}{\mathbf{e}}

\newcommand{\boldb}{\mathbf{b}}
\newcommand{\boldB}{\mathbf{B}}
\newcommand{\boldh}{\mathbf{h}}

\newcommand{\E}{\mathbb{E}}
\newcommand{\reals}{\mathbb{R}}

\newenvironment{itemizesquish}{\begin{list}{\labelitemi}{\setlength{\itemsep}{0em}\setlength{\labelwidth}{0.5em}\setlength{\leftmargin}{\labelwidth}\addtolength{\leftmargin}{\labelsep}}}{\end{list}}

\DeclareMathOperator*{\KL}{KL}
\DeclareMathOperator*{\ELBO}{ELBO}

\DeclareMathOperator*{\pop}{pop}
\DeclareMathOperator*{\push}{push}
\DeclareMathOperator*{\Top}{top}

\DeclareMathOperator*{\RELU}{ReLU}
\DeclareMathOperator*{\MLP}{MLP}
\DeclareMathOperator*{\LSTM}{LSTM}

\DeclareMathOperator*{\TreeLSTM}{TreeLSTM}
\DeclareMathOperator*{\softmax}{softmax}

\DeclareMathOperator*{\bern}{Bernoulli}
\DeclareMathOperator*{\cat}{Cat}

\newcommand{\given}{\,|\,}
\newcommand{\param}{;}
\newcommand\blfootnote[1]{%
  \begingroup
  \renewcommand\thefootnote{}\footnote{#1}%
  \addtocounter{footnote}{-1}%
  \endgroup
}
\aclfinalcopy 

\newcommand{\bDiamond}{\mathbin{\Diamond}}

\makeatletter
\newcommand\bigDiamond{\mathop{\mathpalette\bigDi@mond\relax}}
\newcommand\bigDi@mond[2]{%
  \vcenter{\hbox{\m@th
    \scalebox{\ifx#1\displaystyle 2\else1.2\fi}{$#1\Diamond$}%
  }}%
}
\newcommand\bigLozenge{\mathop{\mathpalette\bigL@zenge\relax}}
\newcommand\bigL@zenge[2]{%
  \vcenter{\hbox{\m@th
    \scalebox{\ifx#1\displaystyle 2\else1.2\fi}{$#1\blacklozenge$}%
  }}%
}
\makeatother

\title{Unsupervised Recurrent Neural Network Grammars}

\author{Yoon Kim$^\dag$ ~ Alexander M. Rush$^\dag$  ~ Lei Yu$^{\bDiamond}$  \\  \textbf{Adhiguna Kuncoro}$^{\ddagger,\bDiamond}$ ~  \textbf{Chris Dyer}$^{\bDiamond}$ ~ \textbf{G\'{a}bor Melis}$^{\bDiamond}$ \\ 
\\
 $^\dag$Harvard University ~
$^{\ddagger}$University of Oxford ~
$^{\bDiamond}$DeepMind \\
 {\small \tt \{yoonkim,srush\}@seas.harvard.edu ~  \tt \{leiyu,akuncoro,cdyer,melisgl\}@google.com}}
\date{}

\begin{document}
\maketitle
\begin{abstract}
  Recurrent neural network grammars (RNNG) are generative models of
  language which jointly model syntax and surface structure by
  incrementally generating a syntax tree and sentence in a top-down,
  left-to-right order. Supervised RNNGs achieve strong language
  modeling and parsing performance, but require an annotated corpus of
  parse trees. In this work, we experiment with unsupervised learning
  of RNNGs. Since directly marginalizing over the space of latent
  trees is intractable, we instead apply amortized variational
  inference. To maximize the evidence lower bound, we develop an
  inference network parameterized as a neural CRF constituency parser.
  On language modeling, unsupervised RNNGs perform as well their
  supervised counterparts on benchmarks in English and Chinese. On
  constituency grammar induction, they are competitive with
  recent neural language models that induce tree structures
  from words through attention mechanisms. \vspace{-2mm}
\end{abstract}

\section{Introduction}
\vspace{-2mm}

\blfootnote{\noindent \hspace{-6mm} Work done while  the first author was an intern at DeepMind.  Code available at \url{https://github.com/harvardnlp/urnng}}
Recurrent neural network grammars (RNNGs) \cite{dyer2016rnng} model
sentences by first generating a nested, hierarchical syntactic
structure which is used to construct a context representation to be conditioned upon for upcoming words. Supervised RNNGs
have been shown to outperform standard sequential language models, achieve excellent results on parsing
\cite{dyer2016rnng,kuncoro2017rnng}, better encode syntactic
properties of language \cite{kuncoro2018syntax}, and correlate with
electrophysiological responses in the human brain
\cite{hale2018human}. However, these all require annotated syntactic trees
for training. In this work, we explore unsupervised learning of
recurrent neural network grammars for language modeling and grammar
induction.

The standard setup for unsupervised structure learning is to define a
generative model $p_\theta(\boldx, \boldz)$ over observed data
$\boldx$ (e.g. sentence) and unobserved structure $\boldz$ (e.g. parse
tree, part-of-speech sequence), and maximize the log marginal
likelihood
$\log p_\theta(\boldx) = \log \sum_{\boldz}
p_\theta(\boldx, \boldz)$.
Successful approaches to unsupervised parsing have made strong
conditional independence assumptions (e.g. context-freeness) and
employed auxiliary objectives \cite{klein2002ccm} or priors
\cite{johnson2007pcfg}. These strategies imbue the
learning process with inductive biases that guide the model to discover 
meaningful structures while allowing tractable
algorithms for marginalization; however, they come at the
expense of language modeling performance, particularly compared to
sequential neural models that make no independence assumptions.

Like RNN language models, RNNGs make no independence
assumptions. Instead they encode structural bias
through operations that compose linguistic constituents.  The lack of
independence assumptions contributes to the strong language
modeling performance of RNNGs, but make unsupervised learning challenging. 
First, marginalization is intractable. Second, the biases
imposed by the RNNG are relatively weak compared to those imposed by models
like PCFGs. There is little pressure for non-trivial tree structure to emerge during unsupervised RNNG (URNNG) learning.

In this work, we explore a technique for handling intractable
marginalization while also injecting inductive bias.  Specifically we
employ amortized variational inference \cite{kingma2014vae,rezende2014vae,mnih2014neural} with a
\emph{structured} inference network. Variational inference lets us
tractably optimize a lower bound on the log marginal
likelihood, while employing a structured inference
network encourages non-trivial structure. In particular, a conditional random
field (CRF) constituency parser \cite{finkel2008crf,durrett2015crf},
which makes significant independence assumptions, acts as a guide on
the generative model to learn meaningful trees through regularizing the posterior \cite{ganchev2010post}.

We experiment with URNNGs on English and Chinese and observe that they
perform well as language models compared to their supervised counterparts
and standard neural LMs. In terms of grammar induction, they are
competitive with recently-proposed neural architectures that discover
tree-like structures through gated attention
\cite{shen2018nlm}. Our results, along with other recent work on
joint language modeling/structure learning with deep networks
\cite{shen2018nlm,shen2019ordered,wiseman2018templates,kawakami2018segmental},
suggest that it is possible to learn generative models of language that
model the underlying data well (i.e. assign high likelihood to held-out data) 
and at the same time induce meaningful linguistic structure.

\vspace{-2mm}
\section{Unsupervised Recurrent Neural Network Grammars}
\vspace{-2mm}

\begin{figure}[t]
    \centering
    \includegraphics[scale=0.24]{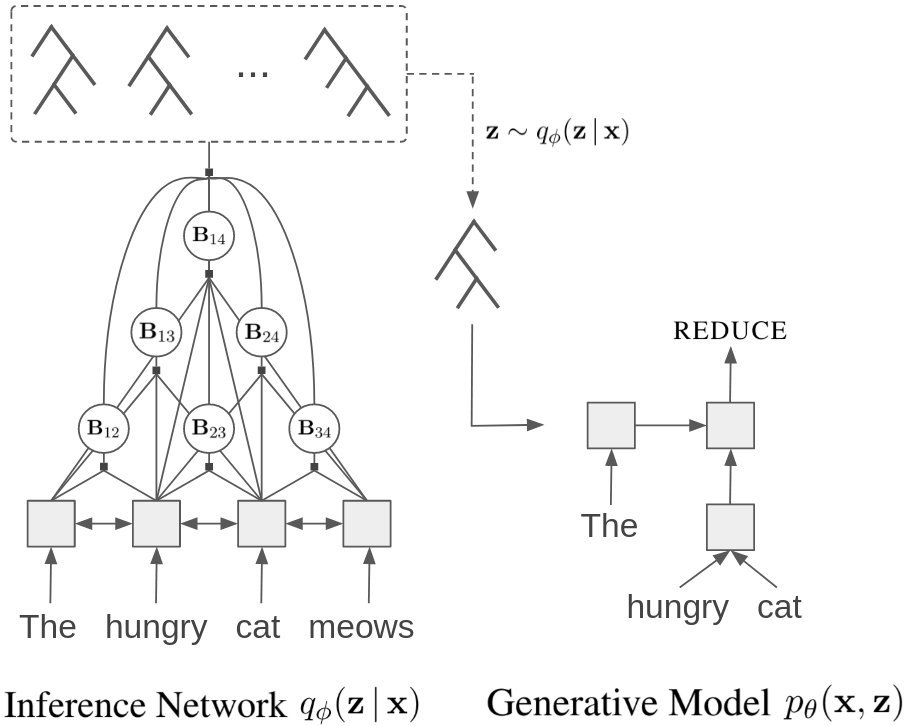}
    \vspace{-6mm}
    \caption{Overview of our approach. The inference network $q_\phi(\boldz \given \boldx)$ (left) is a CRF parser which produces a distribution over binary trees (shown in dotted box). $\boldB_{ij}$ are random variables for existence of a constituent spanning $i$-th and $j$-th words, whose potentials are the output from a bidirectional LSTM (the global factor ensures that the distribution is only over valid binary trees).
    The generative model $p_\theta(\boldx, \boldz)$ (right) is an RNNG which consists of a stack LSTM (from which actions/words are predicted) and a tree LSTM (to obtain constituent representations upon \textsc{reduce}).
    Training involves sampling a binary tree from $q_\phi(\boldz \given \boldx)$, converting it to a sequence of $\text{shift}/\text{reduce}$ actions ($\boldz = $ [\textsc{shift}, \textsc{shift}, \textsc{shift}, \textsc{reduce}, \textsc{reduce}, \textsc{shift}, \textsc{reduce}] in the above example), and optimizing the log joint likelihood $\log p_\theta(\boldx, \boldz)$.}
    \label{fig:urnng}
    \vspace{-2mm}
\end{figure}

We use $\boldx = [x_1, \dots, x_T]$ to denote a sentence of length $T$, and $\boldz \in \mathcal{Z}_T$ to denote an unlabeled binary parse tree over a sequence of length $T$, represented as a binary vector of length $2T-1$. Here 0 and 1 correspond to \textsc{shift} and \textsc{reduce} actions, explained below.\footnote{The cardinality of $\mathcal{Z}_T \subset \{0,1\}^{2T-1}$ is given by the $(T-1)$-th Catalan number, $|\mathcal{Z}_T| = \frac{(2T-2)!}{T!(T-1)!}$. } Figure~\ref{fig:urnng} presents an overview of our approach.
\vspace{-1mm}
\subsection{Generative Model}
\vspace{-1mm}

An RNNG defines a joint probability distribution
$p_\theta(\boldx, \boldz)$ over sentences $\boldx$ and parse trees
$\boldz$. We consider a simplified version of the original RNNG
\cite{dyer2016rnng} by ignoring constituent labels and only
considering binary trees. The RNNG utilizes an RNN to parameterize a
stack data structure \cite{dyer2015stacklstm} of partially-completed
constituents to incrementally build the parse tree while generating
terminals. Using the current stack representation, the model samples
an action (\textsc{shift} or \textsc{reduce}): $\textsc{shift}$
generates a terminal symbol, i.e. word, and shifts it onto the stack,\footnote{A
  better name for $\textsc{shift}$ would be $\textsc{generate}$ (as in
  \citet{dyer2016rnng}), but we use $\textsc{shift}$ to emphasize
  similarity with the shift-reduce parsing.} $\textsc{reduce}$ pops
the last two elements off the stack, composes them, and shifts the
composed representation onto the stack.

Formally, let $S = [(\boldsymbol{0}, \boldsymbol{0})]$ be the initial stack. Each item of the stack will be a pair, where the first element is the hidden state of the stack LSTM, and the second element is an input vector, described below. We use $\Top(S)$ to refer to the top pair in the stack. The $\push$ and $\pop$ operations are defined imperatively in the usual way. 
At each time step, the next action $z_t$ (\textsc{shift} or \textsc{reduce}) is sampled from a Bernoulli distribution parameterized in terms of the current stack representation. Letting $(\boldh_{\text{prev}}, \boldg_{\text{prev}}) = \Top(S)$, we have
        \vspace{-1mm}
    \begin{equation*}
        z_t \sim \bern(p_t), \,\,\,\,\,\,\,p_t = \sigma(\boldw^\top \boldh_{\text{prev}} + b).   \vspace{-1mm}
    \end{equation*}
Subsequent generation depend on $z_t$:
\vspace{-1mm}
    \begin{itemizesquish}
    \item  If $z_t = 0$ (\textsc{shift}), the model first generates a terminal symbol via sampling from a categorical distribution whose parameters come from an affine transformation and a softmax,
    \vspace{-1mm}
        \begin{equation*}
             x\sim \softmax(\boldW \boldh_{\text{prev}} + \boldb). \vspace{-1mm}
        \end{equation*}
        Then the generated terminal is shifted onto the stack using a stack LSTM,
        \vspace{-1mm}
        \begin{align*}
             &\boldh_\text{next} = \LSTM(\bolde_{x}, \boldh_\text{prev}), \\
             &\push(S, (\boldh_\text{next}, \bolde_{x})),
        \end{align*}
        where $\bolde_{x}$ is the word embedding for $x$.
        \vspace{-2mm}
    \item If $z_t = 1$ (\textsc{reduce}), we pop the last two elements off the stack,
            \vspace{-1.5mm}
        \begin{align*}
            (\boldh_{r}, \boldg_{r}) = \pop(S),\,\,\,\,\,  (\boldh_{l}, \boldg_{l}) = \pop(S),
        \end{align*}
        and obtain a new representation that combines the left/right constituent representations using a tree LSTM \cite{tai2015treelstm,zhu2015treelstm},
            \vspace{-2mm}
        \begin{equation*}
            \boldg_{\text{new}} = \TreeLSTM(\boldg_l, \boldg_r).\vspace{-1mm}
        \end{equation*}
        Note that we use $\boldg_l$ and $\boldg_r$ to obtain the new representation instead of $\boldh_l$ and $\boldh_r$.\footnote{The update equations for the tree LSTM (and the stack LSTM) also involve \emph{cell} states in addition to the hidden states. To reduce notational clutter we do not explicitly show the cell states and instead subsume them into $\boldg$. If one (or both) of the inputs to the tree LSTM is a word embedding, the associated cell state is taken to be zero. See \citet{tai2015treelstm} for the exact parameterization.}
        We then update the stack using $\boldg_{\text{new}}$,
                \vspace{-1mm}
        \begin{align*}
        &(\boldh_{\text{prev}}, \boldg_{\text{prev}}) = \Top(S),  \\
        &\boldh_\text{new} = \LSTM(\boldg_\text{new}, \boldh_\text{prev}), \\
        &\push(S, (\boldh_\text{new}, \boldg_\text{new})).
        \end{align*}
\end{itemizesquish}
\vspace{-2mm}
The generation process continues until an end-of-sentence symbol is generated.
The parameters $\theta$ of the generative model are $\boldw, b, \boldW, \boldb$, and the parameters of the stack/tree LSTMs.
For a sentence $\boldx = [x_1, \dots, x_T]$ of length $T$, the binary parse tree is given by the binary vector $\boldz = [z_1, \dots, z_{2T-1}]$.\footnote{As it stands, the support of $\boldz$ is $\{0, 1\}^{2T-1}$, all binary vectors of length $2T-1$. To restrict our distribution to $\mathcal{Z}_T$ (binary vectors which describe valid trees), we constrain $z_t$ to be valid at each time step, which amounts to deterministically choosing $z_t = 0$ (\textsc{shift}) if there are fewer than two elements (not counting the initial zero tuple) on the stack. }
The joint log likelihood decomposes as a sum of terminal/action log likelihoods,
        \vspace{-1mm}
\begin{align*}
    \log p_\theta(&\boldx, \boldz) = \underbrace{\sum_{t=1}^T \log p_\theta(x_t \given \boldx_{<t}, \boldz_{<n(t)})}_{\log p_\theta(\boldx \given \boldz)}\\&
    + \underbrace{\sum_{j=1}^{2T-1} \log p_\theta(z_j \given \boldx_{<m(j)}, \boldz_{<j})}_{\log p_\theta(\boldz \given \boldx_{<\boldz})}, \numberthis \label{eq:joint}
\end{align*}
where $\boldz_{<n(t)}$ refers to all actions before generating the $t$-th word, and similarly $\bold{x}_{<m(j)}$ refers to all words generated before taking the $j$-th action. For brevity, from here on we will use $\log p_\theta(\boldx \given \boldz)$ to refer to the first term (terminal log likelihood) and $\log p_\theta(\boldz \given \boldx_{<\boldz})$ to
refer to the second term (action log likelihood) in the above decomposition.\footnote{The action log likelihood is the sum of log \emph{conditional} priors, which is obviously different from the unconditional log prior $\log p_\theta(\boldz) = \log \sum_{\boldx} p_\theta(\boldx, \boldz)$.}

In the supervised case where ground-truth $\boldz$ is available, we can straightforwardly perform gradient-based optimization to maximize the joint log likelihood $\log p_\theta(\boldx, \boldz)$. In the unsupervised case, the standard approach is to maximize the log marginal likelihood,
\vspace{-1mm}
\begin{equation*}
    \log p_\theta(\boldx) = \log \sum_{\boldz' \in \mathcal{Z}_T} p_\theta(\boldx, \boldz'). \vspace{-1mm}
\end{equation*}
However this summation is intractable because $z_t$ fully depends on all previous actions $[z_1, \dots, z_{t-1}]$. Even if this summation were tractable, it is not clear that
meaningful latent structures would emerge given the lack of explicit independence assumptions in the RNNG (e.g. it is clearly not context-free).
We handle these issues with amortized variational inference.

\vspace{-1mm}
\subsection{Amortized Variational Inference}
\vspace{-1mm}
Amortized variational inference  \cite{kingma2014vae} defines a trainable inference network $\phi$ that parameterizes $q_\phi(\boldz \given \boldx)$, a variational posterior distribution, in this case over parse trees $\boldz$ given the sentence $\boldx$. This distribution is used to form an  evidence lower bound (ELBO) on the log marginal likelihood,
\vspace{-1.5mm}
\begin{equation*} \ELBO(\theta, \phi \param \boldx) = \E_{q_\phi(\boldz \given \boldx)}\left[\log \frac{p_\theta(\boldx, \boldz)}{q_\phi(\boldz \given \boldx)} \right]. \vspace{-1.5mm}
\end{equation*}
We maximize the ELBO with respect to both model parameters $\theta$ and inference network parameters $\phi$.  The
ELBO is still intractable to calculate exactly, but this formulation
will allow us to obtain unbiased gradient estimators based on Monte
Carlo sampling.

Observe that rearranging the ELBO gives the following optimization problem,
\vspace{-1.5mm}
\begin{equation*}
    \max_{\theta, \phi}\,\, \log p_\theta(\boldx) - \KL[q_\phi(\boldz \given \boldx)\, \Vert \, p_\theta(\boldz \given \boldx)].\vspace{-2mm}
\end{equation*} Thus, $\phi$ is trained to match the variational posterior $q_\phi(\boldz \given \boldx)$ to the true posterior $p_\theta(\boldz \given \boldx)$, but $\theta$ is \emph{also} trained to match the true posterior to the variational posterior. Indeed, there is some evidence to suggest that generative models trained with amortized variational inference (i.e. variational autoencoders) learn posterior distributions that are close to the variational family \cite{Cremer2018}.

We can use this to our advantage with an inference network that
injects inductive bias.  We propose to do this by using a context-free
model for the inference network, in particular, a neural CRF parser
\cite{durrett2015crf}. This choice can seen as a form of
posterior regularization that limits posterior flexibility of the overly powerful  RNNG generative model.\footnote{While it has a similar goal, 
  this formulation differs the from posterior
  regularization as formulated by \citet{ganchev2010post}, which constrains
  the distributional family via linear constraints on
  posterior expectations. In our case, the conditional independence assumptions
  in the CRF lead to a \emph{curved} exponential family where the vector of natural parameters 
  has fewer dimensions than the vector of  sufficient statistics of
  the full exponential family. This curved exponential family is 
  a subset of the marginal polytope of the full exponential family, but
  it is an intersection of both linear and nonlinear manifolds, and
  therefore cannot be characterized through linear
  constraints over posterior expectations.}$^{,}$\footnote{In preliminary experiments, we also
  attempted to learn latent trees with a transition-based parser
  (which does not make explicit independence assumptions) that looks
  at the entire sentence. However we found that under this setup, the
  inference network degenerated into a local minimum whereby it always
  generated left-branching trees despite various optimization
  strategies. \citet{williams2018latent} observe a similar phenomenon
  in the context of learning latent trees for classification
  tasks. However \citet{li2019grammar} find that it is possible use a
  transition-based parser as the inference network for dependency
  grammar induction, if the inference network is constrained via
  posterior regularization \cite{ganchev2010post} based on universal
  syntactic rules \cite{naseem2010univ}.}

 The parameterization of span scores is similar to recent works \cite{wang2016graph,stern2017minimal,kitaev2018parsing}: we add position embeddings to word embeddings and run a bidirectional LSTM over the input representations to obtain the forward $[\overrightarrow{\boldh}_1, \dots, \overrightarrow{\boldh}_T]$ and backward $[\overleftarrow{\boldh}_1, \dots, \overleftarrow{\boldh}_T]$ hidden states. The score $s_{ij}\in \reals$ for a constituent spanning $x_i$ to $x_j$ is given by,
\vspace{-1mm}
\begin{equation*}
 s_{ij} = \MLP([\overrightarrow{\boldh}_{j+1} - \overrightarrow{\boldh}_{i}; \overleftarrow{\boldh}_{i-1} - \overleftarrow{\boldh}_{j}]). \vspace{-1mm}
\end{equation*}
Letting $\boldB$ be the binary matrix
representation of a tree ($\boldB_{ij}=1$ means there is a constituent spanning $x_i$ and $x_j$), the CRF parser defines a distribution over binary trees via the Gibbs distribution,
\vspace{-1mm}
\begin{equation*}
q_\phi( \boldB \given \boldx) = \frac{1}{Z_T(\boldx)}\exp \Big(\sum_{i\le j} \boldB_{ij} s_{ij}\Big),      \vspace{-1mm}
\end{equation*}
where  $Z_T(\boldx) $ is the partition function,
\vspace{-1mm}
\begin{equation*}
    Z_T(\boldx) = \sum_{\boldB' \in \mathcal{B}_T} \exp \Big(\sum_{i\le j} \boldB_{ij}' s_{ij}\Big), \vspace{-1mm}
\end{equation*}
and $\phi$ denotes the parameters of the inference network (i.e. the bidirectional LSTM and the MLP). Calculating $Z_T(\boldx)$ requires a summation over an exponentially-sized set $\mathcal{B}_T \subset \{0,1 \}^{T \times T}$, the set of all binary trees over a length $T$ sequence. However we can perform the summation in $O(T^3)$ using the inside algorithm \cite{baker1979io}, shown in Algorithm~\ref{alg:inside}. This computation is itself differentiable and amenable to gradient-based optimization.
Finally, letting $f : \mathcal{B}_T \rightarrow \mathcal{Z}_T$ be the bijection between the  binary tree matrix representation  and a sequence of \textsc{shift}/\textsc{reduce} actions, the inference network defines a distribution over $\mathcal{Z}_T$ via  $q_\phi(\boldz \given \boldx) \triangleq q_\phi( f^{-1}(\boldz) \given \boldx)$.
\begin{algorithm}[t]
    \caption{Inside algorithm for calculating $Z_T(\boldx)$}
    \label{alg:inside}
    \begin{algorithmic}[1] 
      \Procedure{Inside}{$s$} \Comment{scores $s_{ij}$ for $i\le j$}
      \For{$i:=1$ \textbf{to} $T$} \Comment{length-1 spans}
                    \State $\beta[i,i] = \exp(s_{ii})$
                    \EndFor
            \For{$\ell:=1$ \textbf{to} $T-1$} \Comment{span length}
            \For{$i:=1$ \textbf{to} $T-\ell$} \Comment{span start}
            \State $j = i + \ell$ \Comment{span end}
            \State $\beta[i,j] = \sum_{k=i}^{j-1}\exp(s_{ij})\cdot\beta[i,k]\cdot\beta[k+1,j]$
            \EndFor
            \EndFor
            \State \textbf{return} $\beta[1,T]$\Comment{return partition function $Z_T(\boldx)$}
        \EndProcedure
    \end{algorithmic}
\end{algorithm}

\vspace{-2mm}
\subsection{Optimization}
\vspace{-1mm}
\label{sec:opt}
For optimization, we use the following variant of the ELBO, 
\vspace{-2mm}
\begin{equation*}
 \E_{q_\phi(\boldz \given \boldx)}[\log p_\theta(\boldx, \boldz)] + \mathbb{H}[q_\phi(\boldz \given \boldx)],
\end{equation*}
where $\mathbb{H}[q_\phi(\boldz \given \boldx)] = \E_{q_\phi(\boldz \given \boldx)}[- \log q_\phi(\boldz \given \boldx)]$ is the entropy of the variational posterior.
A Monte Carlo estimate for the gradient with respect to $\theta$ is
\vspace{-2mm}
\begin{equation*}
    \nabla_\theta \ELBO(\theta, \phi \param \boldx) \approx \frac{1}{K} \sum_{k=1}^{K} \nabla_\theta \log p_\theta(\boldx, \boldz^{(k)}), \vspace{-1mm}
\end{equation*}
with samples $\boldz^{(1)}, \dots, \boldz^{(K)}$ from
$q_\phi(\boldz \given \boldx)$. Sampling uses the intermediate values
calculated during the inside algorithm to sample split points
recursively \cite{goodman1998parsing,finkel2006}, 
as shown in Algorithm~\ref{alg:sample}.  The gradient with
respect to $\phi$ involves two parts. The entropy term
$\mathbb{H}[q_\phi(\boldz \given \boldx)]$ can be calculated exactly
in $O(T^3)$, again using the intermediate values from the inside
algorithm (see Algorithm~\ref{alg:entropy}).\footnote{We adapt the algorithm for calculating tree entropy in PCFGs from \citet{hwa2000entropy} to the CRF case.} Since each step of this dynamic program is
differentiable, we can obtain the gradient
$\nabla_\phi \mathbb{H}[q_\phi(\boldz \given \boldx)]$ using
automatic differentation.\footnote{$\nabla_\phi \mathbb{H}[q_\phi(\boldz \given \boldx)]$ can also be computed using the inside-outside algorithm and a second-order expectation semiring~\cite{li:2009}, which has the same asymptotic runtime complexity but generally better constants.} An estimator for the gradient with
respect to
$\E_{q_\phi(\boldz \given \boldx)}[\log p_\theta(\boldx, \boldz)]$ is
obtained via the score function gradient estimator
\cite{glynn1987ratio,williams1992reinforce}, \vspace{-1.5mm}
\begin{align*}
     \nabla_\phi &\E_{q_\phi(\boldz \given \boldx)}[\log p_\theta(\boldx, \boldz)] \\ &=  \E_{q_\phi(\boldz \given \boldx)}[\log p_\theta(\boldx, \boldz)\nabla_\phi \log q_\phi(\boldz \given \boldx)] \\
     &\approx \frac{1}{K} \sum_{k=1}^K \log p_\theta(\boldx, \boldz^{(k)})\nabla_\phi\log q_\phi(\boldz^{(k)} \given \boldx).
\end{align*}
The above estimator is unbiased but typically suffers from high variance.
To reduce variance, we use a control variate derived from an average of the other samples' joint likelihoods \cite{mnih2016vimco}, yielding the following estimator,
\vspace{-1.5mm}
\begin{equation*}
    \frac{1}{K} \sum_{k=1}^K (\log p_\theta(\boldx, \boldz^{(k)}) - r^{(k)})\nabla_\phi\log q_\phi(\boldz^{(k)} \given \boldx), \vspace{-1.5mm}
\end{equation*}
where $r^{(k)} = \frac{1}{K-1} \sum_{j \ne k} \log p_\theta(\boldx, \boldz^{(j)})$.
This control variate worked better than alternatives such as estimates of
baselines from an auxiliary network \cite{mnih2014neural,deng2018} or a language model \cite{yin2018structvae}.

\begin{algorithm}[t]
    \caption{Top-down sampling a tree from $q_\phi(\boldz \given \boldx)$}
    \label{alg:sample}
    \begin{algorithmic}[1] 
        \Procedure{Sample}{$\beta$} \Comment{$\beta$ from running  \textsc{Inside}(s)}
        \State $\boldB = \mathbf{0}$ \Comment{binary matrix representation of tree}
        \State $Q = [(1, T)]$ \Comment{queue of constituents}
        \While{$Q$ is not empty}
        \State $(i, j) = \pop(Q)$
        \State $\tau = \sum_{k=i}^{j-1} \beta[i,k] \cdot \beta[k+1,j]$
            \For{$k:=i$ to  $j-1$} \Comment{get distribution over splits}
            \State $w_{k} = (\beta[i,k] \cdot \beta[k+1,j]) / \tau $
                     \EndFor
        \State $k \sim \cat([w_i, \dots, w_{j-1}])$ \Comment{sample a split point}
        \State $\boldB_{i, k} =1,\  \boldB_{k+1,j} = 1 $ \Comment{update $\boldB$}
        \If{$k > i$} \Comment{if left child has width $>$ 1 }
        \State $\push(Q, (i, k))$ \Comment{add to queue}
                \EndIf
        \If{$k +1 < j$} \Comment{if right child has width $>$ 1 }
        \State $\push(Q, (k+1, j))$  \Comment{add to queue}
        \EndIf
            \EndWhile
            \State $\boldz = f(\boldB)$ \Comment{\parbox[t]{.67\linewidth}{$f:\mathcal{B}_T \to \mathcal{Z}_T$ maps matrix representation of tree to sequence of actions.}}
            \State \textbf{return} $\boldz$
        \EndProcedure
    \end{algorithmic}
\end{algorithm}

\vspace{-2mm}
\section{Experimental Setup}
\vspace{-2mm}
\subsection{Data}\vspace{-1mm} \label{sec:data}
For English we use the Penn Treebank \cite[PTB]{marcus1993ptb} with splits and preprocessing from \citet{dyer2016rnng} which retains punctuation and replaces singleton words with Berkeley parser's mapping rules, resulting in a vocabulary of 23,815 word types.\footnote{\url{https://github.com/clab/rnng}} Notably this is much larger than the standard PTB LM setup from \citet{Mikolov2010} which uses 10K types.\footnote{Both versions of the PTB data can be obtained from  \url{http://demo.clab.cs.cmu.edu/cdyer/ptb-lm.tar.gz}.}
Also different from the LM setup, we model each sentence separately instead of carrying information across sentence boundaries, as the RNNG is a generative model of sentences. Hence our perplexity numbers are not comparable to the PTB LM results \cite{melis2018sota,merity2018reg,yang2018softmax}. 

Since the PTB is rather small, and since the URNNG does not require annotation, we also test our approach on a subset of the one billion word corpus \cite{chelba2013oneb}. We randomly sample 1M sentences for training and 2K sentences for validation/test, and limit the vocabulary to 30K word types. While still a subset of the full corpus (which has 30M sentences), this dataset is two orders of magnitude larger than PTB.
Experiments on Chinese utilize version 5.1 of the Chinese Penn Treebank (CTB) \cite{xue2005ctb}, with
the same splits as in \citet{chen2014fast}. Singleton words are replaced with a single \textsc{$\langle$unk$\rangle$} token, resulting in a vocabulary of 17,489 word types.
\vspace{-1mm}
\begin{algorithm}[t]
    \caption{Calculating the tree entropy $\mathbb{H}[q_\phi(\boldz \given \boldx)]$}
    \label{alg:entropy}
    \begin{algorithmic}[1] 
        \Procedure{Entropy}{$\beta$}  \Comment{$\beta$ from running  \textsc{Inside}(s)}
        \For{$i:=1$ to $T$} \Comment{initialize entropy table}
                    \State $H[i,i] =0$
                    \EndFor
            \For{$l:=1$ to $T-1$} \Comment{span length}
            \For{$i:=1$ to $T-l$} \Comment{span start}
            \State $j = i + l$ \Comment{span end}
            \State $\tau = \sum_{u=i}^{j-1} \beta[i,u] \cdot \beta[u+1,j]$
            \For{$u:=i$ to  $j-1$}
            \State $w_{u} = (\beta[i,u] \cdot \beta[u+1,j]) / \tau $
                     \EndFor
            \State $H[i,j] =\sum_{u=i}^{j-1}(H[i,u]+H[u+1,j] $
            \State $\,\,\,\,\,\,\,\,\,\,\,\,\,\,\,\,\,\,\,\,\,\,\,\, - \log w_u)\cdot w_u$
            \EndFor
            \EndFor
            \State \textbf{return} $H[1,T]$\Comment{return tree entropy $\mathbb{H}[q_\phi(\boldz \given \boldx)]$}
        \EndProcedure
    \end{algorithmic}
\end{algorithm}
\vspace{-1mm}
\subsection{Training and Hyperparameters}
\vspace{-1mm}
The stack LSTM has two layers with input/hidden size equal to 650 and dropout of 0.5. The tree LSTM also has 650 units. The inference network uses a one-layer bidirectional LSTM with 256 hidden units, and the MLP (to produce span scores $s_{ij}$ for $i \le j$) has a single hidden layer with a $\RELU$ nonlinearity followed by layer normalization \cite{ba2016layernorm} and dropout of 0.5. We share word embeddings between the generative model and the inference network, and also tie weights between the input/output word embeddings \cite{press2016tie}.

Optimization of the model itself required standard techniques for avoiding posterior collapse in VAEs.\footnote{Posterior collapse in our context means that $q_\phi(\boldz \given \boldx)$ always produced trivial (always left or right branching) trees.} We warm-up the ELBO objective by linearly annealing (per batch) the weight on the conditional prior $\log p_\theta(\boldz \given \boldx_{<\boldz})$ and the entropy $\mathbb{H}[q_\phi(\boldz \given \boldx)]$ from 0 to 1 over the first two epochs (see equation (\ref{eq:joint}) for definition of $\log p_\theta(\boldz \given \boldx_{<\boldz})$). This is analogous to KL-annealing in
VAEs with continuous latent variables \cite{bowman2016vae,Son2016}.
We train for 18 epochs (enough for convergence for all models) with a batch size of 16 and $K=8$ samples for the Monte Carlo gradient estimators.
The generative model is optimized with SGD with learning rate equal to 1, except for the affine layer that produces a distribution over the actions, which has learning rate 0.1. Gradients of the generative model are clipped at 5. 
The inference network is optimized with Adam \cite{kingma2015adam} with learning rate 0.0001, $\beta_1 =0.9, \beta_2 = 0.999$, and gradient clipping at 1. As Adam converges significantly faster than SGD (even with a much lower learning rate), we stop training the inference network after the first two epochs. 
Initial model parameters are sampled from $\mathcal{U}[-0.1, 0.1]$.
The learning rate starts decaying by a
factor of 2 each epoch after the first epoch at which validation
performance does not improve, but this learning rate
decay is not triggered for the first eight epochs to ensure adequate training. 
We use the same hyperparameters/training setup for both PTB and CTB. For experiments on (the subset of) the one billion word corpus, we use a smaller dropout rate of 0.1. The baseline RNNLM also uses the smaller dropout rate.

All models are trained with an end-of-sentence token, but for perplexity calculation these tokens  are not counted to be comparable to prior work \cite{dyer2016rnng,kuncoro2017rnng,buys2018syntax}. To be more precise, the inference network does not make use of the end-of-sentence token to produce
parse trees, but the generative model is trained to generate the end-of-sentence token
after the final \textsc{reduce} operation.
\vspace{-2mm}
\subsection{Baselines}{\label{sec:baselines}}
\vspace{-1mm}
We compare the unsupervised RNNG (URNNG) against several baselines: (1) RNNLM, a standard RNN language model
whose size is the same as URNNG's stack LSTM; (2) Parsing Reading Predict Network (PRPN)
\cite{shen2018nlm}, a neural language model that uses gated attention layers to embed soft tree-like structures into a neural network (and among the current state-of-the-art in grammar induction from words on the full corpus); (3) RNNG with trivial trees (left branching, right branching, random); (4) supervised RNNG trained on unlabeled, binarized gold trees.\footnote{We use right branching binarization---\citet{matsu2005pcfg} find that differences between various binarization schemes have marginal impact. Our supervised RNNG therefore differs the original RNNG, which trains on non-binarized trees and does not ignore constituent labels.} Note that the supervised RNNG also trains a discriminative parser
$q_\phi(\boldz \given \boldx)$ (alongside the generative model $p_\theta(\boldx, \boldz)$) in order to sample parse forests for perplexity evaluation (i.e. importance sampling). This discriminative parser has the same architecture as URNNG's inference network. For all models, we perform early stopping based on validation perplexity.
\vspace{-3mm}
\section{Results and Discussion}\label{sec:results}
\vspace{-2mm}
\begin{table}[t]
\small
    \centering
    \begin{tabular}{l r r r r r}
    \toprule
    & \multicolumn{2}{c}{PTB} & & \multicolumn{2}{c}{CTB} \\
    Model & PPL & $F_1$ & & PPL & $F_1$ \\
    \midrule
         RNNLM & 93.2 & -- & & 201.3 & -- \\
         PRPN (default) & 126.2 &32.9 & & 290.9 &32.9 \\
         PRPN (tuned) & 96.7 & 41.2 & & 216.0 & 36.1 \\
        Left Branching Trees &100.9 & 10.3  & & 223.6 &12.4  \\
        Right Branching Trees& 93.3 & 34.8  & &203.5  & 20.6 \\
        Random Trees& 113.2 & 17.0 & &209.1 & 17.4 \\
        URNNG & 90.6 & 40.7 & &195.7 & 29.1\\
        \midrule
        RNNG &  88.7 & 68.1 & &193.1 & 52.3 \\
        RNNG $\to$ URNNG & 85.9 & 67.7  & & 181.1 & 51.9\\
        \midrule
        Oracle Binary Trees & -- & 82.5 & & -- & 88.6 \\
         \bottomrule
    \end{tabular}
    \vspace{-2mm}
    \caption{Language modeling perplexity (PPL) and grammar induction $F_1$ scores on English (PTB) and Chinese (CTB) for the different models. We separate results for those that make do not make use of annotated data (top) versus those that do (mid). Note that our PTB setup from \citet{dyer2016rnng} differs considerably from the usual language modeling setup \cite{Mikolov2010} since we model each sentence independently and use a much larger vocabulary (see \cref{sec:data}).}
    \label{tab:main}
     \vspace{-2mm}   
\end{table}
\vspace{-1mm}
\subsection{Language Modeling}
\vspace{-1mm}
Table~\ref{tab:main} shows perplexity for the different models on PTB/CTB.
As a language model URNNG outperforms an RNNLM and is competitive with the supervised RNNG.\footnote{For RNNG and URNNG we estimate the log marginal likelihood (and hence, perplexity) with $K=1000$ importance-weighted samples,
$ \log p_\theta(\boldx) \approx \log \Big(\frac{1}{K} \sum_{k=1}^K \frac{\log p(\boldx, \boldz^{(k)})}{q_\phi(\boldz^{(k)}\given \boldx)}\Big)$. During evaluation only, we also flatten $q_\phi(\boldz^{}\given \boldx)$ by dividing span scores $s_{ij}$ by a temperature term $2.0$ before feeding it to the CRF.}
The left branching baseline performs poorly, implying that the strong performance of URNNG/RNNG is not simply due to the additional depth afforded by the tree LSTM composition function (a left branching tree, which always performs \textsc{reduce} when possible, is the ``deepest" model). The right branching baseline is essentially equivalent to an RNNLM and hence performs similarly. We found PRPN with default hyperparameters (which obtains a perplexity of 62.0 in the PTB setup from \citet{Mikolov2010}) to not perform well, but tuning hyperparameters improves performance.\footnote{Using the code from \url{https://github.com/yikangshen/PRPN}, we tuned model size, initialization, dropout, learning rate, and use of batch normalization.}
The supervised RNNG performs well as a language model, despite being trained on the joint (rather than marginal) likelihood objective.\footnote{RNNG is trained to maximize $\log p_\theta(\boldx, \boldz)$ while URNNG is trained to maximize (a lower bound on) the  language modeling objective $\log p_\theta(\boldx)$.} This indicates that explicit modeling of syntax helps generalization even with richly-parameterized neural models.
Encouraged by these observations, we also experiment with a hybrid approach where we train a supervised RNNG first and continue fine-tuning the model (including the inference network)
on the URNNG objective (RNNG $\to$ URNNG in Table~\ref{tab:main}).\footnote{We fine-tune for 10 epochs and use a smaller learning rate of 0.1 for the generative model.} This approach results in nontrivial perplexity improvements, and suggests that it is potentially possible to improve language models with supervision on parsed data.
\begin{figure}[t]
    \centering
    \includegraphics[scale=0.40]{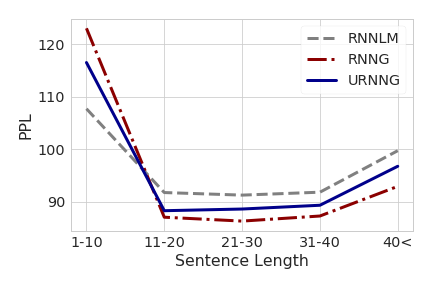}
        \vspace{-5mm}
    \caption{Perplexity of the different  models grouped by sentence length on PTB.}
           \vspace{-2mm}
    \label{fig:len}
\end{figure}
In Figure~\ref{fig:len} we show perplexity by sentence length. We find that a standard language model (RNNLM) is better at modeling short sentences, but underperforms models that explicitly take into account structure (RNNG/URNNG) when the sentence length is greater than 10.
Table~\ref{tab:lm} (top) compares our results against prior work on this version of the PTB, and Table~\ref{tab:lm} (bottom) shows the results on a 1M sentence subset of the one billion word corpus, which is two orders of magnitude larger than PTB. On this larger
dataset URNNG still improves upon the RNNLM. We also trained an RNNG (and RNNG $\to$ URNNG) on this dataset by parsing the training set with the self-attentive parser from \citet{kitaev2018parsing}.\footnote{To parse the training set we use the \texttt{benepar\_en2} model from  \url{https://github.com/nikitakit/self-attentive-parser}, which obtains an $F_1$ score of 95.17 on the PTB test set.} These models improve upon the RNNLM but not the URNNG, potentially highlighting the limitations of using predicted trees for supervising RNNGs.

\begin{table}[t]
    \small
    \centering
    \begin{tabular}{ l r }
    \toprule
     PTB  & PPL \\
    \midrule
    KN 5-gram \cite{dyer2016rnng} & 169.3 \\
    RNNLM  \cite{dyer2016rnng} & 113.4 \\
    Original RNNG   \cite{dyer2016rnng} & 102.4 \\
    Stack-only RNNG   \cite{kuncoro2017rnng} & 101.2 \\
    Gated-Attention RNNG   \cite{kuncoro2017rnng} & 100.9 \\
    Generative Dep. Parser \cite{buys2015gen} & 138.6 \\
    RNNLM \cite{buys2018syntax} & 100.7 \\
    Sup. Syntactic NLM  \cite{buys2018syntax} & 107.6 \\
    Unsup. Syntactic NLM   \cite{buys2018syntax} & 125.2 \\
    PRPN$^\dag$ \cite{shen2018nlm}  & 96.7 \\
    This work: \\
         \hspace{2mm} RNNLM  & 93.2 \\
          \hspace{2mm} URNNG & 90.6\\
         \hspace{2mm} RNNG &  88.7 \\
         \hspace{2mm}  RNNG $\to$ URNNG & 85.9 \\
        \midrule
     1M Sentences & PPL \\
    \midrule
        PRPN$^\dag$ \cite{shen2018nlm}  & 77.7 \\
        RNNLM & 77.4 \\
        URNNG & 71.8 \\
        RNNG$^\ddagger$ & 72.9 \\
        RNNG$^\ddagger$ $\to$ URNNG & 72.0 \\
         \bottomrule
    \end{tabular}
        \vspace{-2mm}
    \caption{(Top) Comparison of this work as a language model against prior works on
    sentence-level PTB with preprocessing from \citet{dyer2016rnng}. Note that previous versions of RNNG differ from ours in terms of parameterization and model size. (Bottom) Results on a subset (1M sentences) of  the one billion word corpus.  PRPN$^\dag$ is
    the model from \citet{shen2018nlm}, whose hyperparameters were tuned by us.  RNNG$^\ddagger$ is trained on predicted parse trees from the self-attentive parser from \citet{kitaev2018parsing}.}
    \label{tab:lm}
\vspace{-2mm}
\end{table}
\vspace{-2mm}
\subsection{Grammar Induction}
\vspace{-1mm}
Table~\ref{tab:main} also shows the $F_1$ scores for grammar induction. Note that we induce latent trees directly from words on the full dataset.\footnote{Past work on grammar induction usually train/evaluate on short sentences and also assume access to gold POS tags \cite{klein2002ccm,smith2004,bod2006subtrees}. However more recent works train directly on words \cite{jin2018depth,shen2018nlm,drozdov2018latent}.}
For RNNG/URNNG we obtain the highest scoring tree from $q_\phi(\boldz \given \boldx)$ through the Viterbi inside (i.e. CKY) algorithm.\footnote{Alternatively, we could estimate $\arg\max_\boldz p_\theta(\boldz \given \boldx)$ by sampling parse trees from $q_\phi(\boldz \given \boldx)$ and using $p_\theta(\boldx, \boldz)$ to rerank the output, as in \citet{dyer2016rnng}.}
We calculate unlabeled $F_1$ using \texttt{evalb}, which
ignores punctuation and discards trivial spans (width-one and sentence spans).\footnote{Available at \url{https://nlp.cs.nyu.edu/evalb/}. We evaluate with  \texttt{COLLINS.prm} parameter file and $\texttt{LABELED}$ option equal to 0. We observe that the setup for grammar induction varies widely across different papers: lexicalized vs. unlexicalized; use of punctuation
vs. not; separation of train/test sets; counting sentence-level spans for evaluation vs. ignoring them; use of additional data; length cutoff for training/evaluation; corpus-level $F_1$ vs. sentence-level $F_1$; and, more. In our survey of twenty or so papers, almost no two papers were identical in their setup. Such variation makes it difficult to meaningfully compare models across papers. Hence, we report grammar induction results mainly for the models and baselines considered in the present work.} Since we compare $F_1$ against the original, non-binarized trees
(per convention), $F_1$ scores of models using oracle binarized trees constitute the upper bounds.

 We confirm the replication study of \citet{htut2018grammar} and find that PRPN is a strong model for grammar induction. URNNG performs on par with PRPN on English but PRPN does better on Chinese; both outperform right branching baselines.
Table~\ref{tab:trees} further analyzes the learned trees and shows the $F_1$ score of URNNG trees against other trees (left), and the recall of URNNG/PRPN trees against ground truth constituents (right). We find that trees induced by URNNG and PRPN are quite different; URNNG is more sensitive to SBAR and VP, while PRPN is better at
identifying NP. While left as future work, this naturally suggests a hybrid approach wherein the intersection of constituents from URNNG and PRPN is used to create a corpus of partially annotated trees, which can be used to guide another model, e.g. via posterior regularization \cite{ganchev2010post} or semi-supervision \cite{hwa1999sup}. 
Finally, Table~\ref{tab:compprior} compares our results using the same evaluation setup as in \citet{drozdov2018latent}, which differs considerably from our setup.
\begin{table}[t]
    \small
    \centering
    \begin{tabular}{l r r}
    \toprule
     Tree & PTB &  CTB \\
    \midrule
Gold & 40.7 & 29.1 \\
Left & 9.2 & 8.4 \\
Right & 68.3 & 51.2 \\
Self & 92.3 & 87.3 \\
RNNG &55.4 & 47.1\\
PRPN & 41.0 & 47.2 \\
         \bottomrule
    \end{tabular}
    \hspace{1mm}
        \begin{tabular}{l r r}
    \toprule
     Label & URNNG &  PRPN \\
    \midrule
SBAR &  74.8$\%$ &  28.9$\%$\\
NP &   39.5$\%$ &63.9$\%$  \\
VP & 76.6$\%$ &  27.3$\%$ \\
PP & 55.8$\%$ & 55.1$\%$\\
ADJP & 33.9$\%$& 42.5$\%$\\
ADVP & 50.4$\%$ & 45.1$\%$ \\
         \bottomrule
    \end{tabular}
        \vspace{-2mm}
    \caption{(Left) $F_1$ scores of URNNG against other trees. ``Self" refers to another URNNG trained with a different random seed. (Right) Recall of constituents
    by label for URNNG and PRPN.  Recall for a particular label is the fraction of ground truth constituents of that label that were identified by the model (as in \citet{htut2018grammar}).}
    \label{tab:trees}
\end{table}
\vspace{-1mm}

\begin{table}[t]
    \small
    \centering
    \vspace{-3mm}
    \begin{tabular}{l r r}
    
    \toprule
    &  $F_1$ &  $+$\textsc{PP} \\
    \midrule
PRPN-UP$^{\ddagger}$ & 39.8 & 45.4 \\
PRPN-LM$^{\ddagger}$ & 42.8 & 42.4 \\
ON-LSTM$^{\ddagger}$  \cite{shen2019ordered} & 49.4 & $-$ \\
DIORA$^{\ddagger}$  \cite{drozdov2018latent} & 49.6 & 56.2 \\
PRPN (tuned) & 49.0 & 49.9 \\
URNNG & 52.4 & 52.4 \\
 \bottomrule
    \end{tabular}
        \vspace{-2mm}
    \caption{PTB $F_1$ scores using the same evaluation setup as \citet{drozdov2018latent}, which evaluates against binarized trees, counts punctuation and trivial spans, and uses sentence-level $F_1$. $+$\textsc{PP} indicates a post-processing heuristic which directly attaches trailing punctuation to the root. This does not change URNNG results since it learns to do so anyway. Results with $^{\ddagger}$ are copied from Table 1 of \citet{drozdov2018latent}. }
    \label{tab:compprior}
          \vspace{-1mm}
\end{table}
\vspace{-1.5mm}
\subsection{Distributional Metrics}
\vspace{-1.5mm}
Table~\ref{tab:entropy} shows some standard metrics related to the learned
generative model/inference network. The ``reconstruction" perplexity
based on $\E_{q_\phi(\boldz \given \boldx)}[\log p_\theta(\boldx \given \boldz)]$ is
much lower than regular perplexity, and further, the Kullback-Leibler divergence between the conditional prior and the variational posterior, given by
\[ \E_{q_\phi(\boldz \given \boldx)} \left[\log \frac{q_\phi(\boldz \given \boldx)}{p_\theta(\boldz \given \boldx_{<\boldz})}\right], \]
is highly nonzero. (See equation (\ref{eq:joint}) for definitions of $\log p_\theta(\boldx \given \boldz)$ and $\log p_\theta(\boldz \given \boldx_{<\boldz})$). This indicates that the latent space is being used in a meaningful way and that there
is no posterior collapse \cite{bowman2016vae}.
As expected, the entropy of the variational posterior is much lower than 
the entropy of the conditional prior, but there is still some uncertainty in the posterior.

\begin{table}[t]
    \small
    \centering
    \begin{tabular}{l r r r r }
    \toprule
    & \multicolumn{2}{c}{PTB} & \multicolumn{2}{c}{CTB} \\
    & RNNG & URNNG & RNNG & URNNG \\
    \midrule
    PPL & 88.7 & 90.6 & 193.1 & 195.7 \\
    Recon. PPL & 74.6 & 73.4 & 183.4 & 151.9 \\
    KL & 7.10 & 6.13  & 11.11 & 8.91 \\
    Prior Entropy & 7.65 & 9.61 & 9.48 & 15.13 \\
    Post. Entropy & 1.56 & 2.28 & 6.23 & 5.75  \\
    Unif. Entropy & 26.07  & 26.07  &  30.17 & 30.17 \\
\bottomrule
    \end{tabular}
        \vspace{-2mm}
    \caption{Metrics related to the generative model/inference network for RNNG/URNNG. For the supervised RNNG we take the ``inference network" to be the discriminative parser trained alongside the generative model (see \cref{sec:baselines}).  Recon. PPL is the reconstruction perplexity based on  $\E_{q_\phi(\boldz \given \boldx)}[\log p_\theta(\boldx \given \boldz)]$, and KL is the Kullback-Leibler divergence. Prior entropy is the entropy of the conditional prior $p_\theta(\boldz \given \boldx_{<\boldz})$, and uniform entropy is the entropy of the uniform distribution over all binary trees. The KL/entropy metrics are averaged across sentences.}
    \label{tab:entropy}
      \vspace{-1mm}
\end{table}
\vspace{-1.5mm}
\subsection{Syntactic Evaluation}
\vspace{-1.5mm}
We perform a syntactic evaluation of the different models based on the setup from
\citet{marvin2018syntax}: the model is given two minimally different sentences, one grammatical and one ungrammatical, and must identify the grammatical sentence by assigning it higher probability.\footnote{We modify the publicly available dataset from \url{https://github.com/BeckyMarvin/LM_syneval} to only keep sentence pairs that did not have any unknown words with respect to our vocabulary, resulting in 80K sentence pairs for evaluation. Further, we train on a much smaller corpus, and hence our results are not directly comparable.} Table~\ref{tab:syntax}
shows the accuracy results.  Overall the supervised RNNG significantly outperforms the other models, indicating opportunities for further work in unsupervised modeling. While the URNNG does slightly outperform an RNNLM, the distribution of errors made from both models are similar, and thus it is not clear whether the outperformance is simply due to better perplexity or learning different structural biases.
\vspace{-1.5mm}
\subsection{Limitations}
\vspace{-1.5mm}
There are several limitations to our approach. For one, the URNNG takes considerably more time/memory to train than a standard language model due to the $O(T^3)$ dynamic program in the inference network, multiple samples to obtain
low-variance gradient estimators, and dynamic computation graphs that make efficient batching nontrivial.\footnote{The main time bottleneck is the dynamic compution graph, since the dynamic programming algorithm can be batched (however the latter is a significant memory bottleneck). We manually batch the \textsc{shift} and \textsc{reduce} operation as much as possible, though recent work on auto-batching \cite{neubig2017batching} could potentially make this  easier/faster.}
The model is sensitive to hyperparameters and required
various optimization strategies (e.g. separate optimizers for the inference network and the generative model) to avoid posterior collapse.
Finally, the URNNG also seemed to rely heavily on punctuation to identify constituents and we were unable to improve upon a right-branching baseline when training the URNNG on a version of PTB where punctuation is removed.\footnote{Many prior works that induce trees directly from words often employ additional heuristics based on punctuation \cite{seginer2007unsup,ponvert2011simple,spitkovsky2013breaking,parikh2014spectral}, as punctuation (e.g. comma) is usually a reliable signal for start/end of constituent spans. The URNNG still has to \emph{learn} to rely on punctuation, similar to recent works such as depth-bounded PCFGs \cite{jin2018depth} and DIORA \cite{drozdov2018latent}. In contrast, PRPN \cite{shen2018nlm} and Ordered Neurons \cite{shen2019ordered} induce trees by directly training on corpus without punctuation. We also reiterate that punctuation is used during training but ignored during evaluation (except in Table~\ref{tab:compprior}).} 
\begin{table}[t]
    \small
    \centering
    \begin{tabular}{l r r r r r}
    \toprule
    & RNNLM & PRPN & RNNG & URNNG \\
    \midrule
 PPL & 93.2 & 96.7 & 88.7 & 90.6 \\
 \midrule
 Overall & 62.5$\%$ & 61.9$\%$& 69.3$\%$ & 64.6$\%$ \\
 $\,\,\,\,$ Subj. & 63.5$\%$ & 63.7$\%$& 89.4$\%$ & 67.2$\%$ \\
 $\,\,\,\,$ Obj. Rel.& 62.6$\%$ & 61.0$\%$& 67.6$\%$ & 65.7$\%$ \\
 $\,\,\,\,$ Refl. & 60.7$\%$ & 68.8$\%$& 57.3$\%$ & 60.5$\%$ \\
 $\,\,\,\,$ NPI & 58.7$\%$ & 39.5$\%$& 46.8$\%$ & 55.0$\%$ \\

 \bottomrule
    \end{tabular}
        \vspace{-2mm}
    \caption{Syntactic evaluation based on the setup from \citet{marvin2018syntax}. Subj. is subject-verb agreement in sentential complement, across prepositional phrase/subjective relative clause, and VP coordination; Obj. Rel. refers to subject-verb agreement in/across an objective relative clause; Refl. refers to reflexive pronoun agreement with antecedent; NPI is negative polarity items.}
    \label{tab:syntax}
          \vspace{-2mm}
\end{table}
\vspace{-3mm}
\section{Related Work}
\vspace{-2mm}
There has been much work on incorporating tree structures into deep models for syntax-aware language modeling, both for unconditional \cite{emami2005nslm,buys2015gen,dyer2016rnng} and conditional \cite{yin2017syntax,melis2017drnn,rabinovich2017asn,aharoni2017tree,eriguchi2017nmt,wang2018treenmt,gu2018topdown} cases. These approaches generally rely on annotated parse trees during training and maximizes the joint
 likelihood of sentence-tree pairs. Prior work
 on combining language modeling and unsupervised tree learning typically
 embed soft, tree-like structures as hidden layers of a deep network \cite{cho2014prop,chung2017hier,shen2018nlm,shen2019ordered}. In contrast, \citet{buys2018syntax} make Markov assumptions and perform exact marginalization over latent dependency trees. Our work is also related to the recent line of work on learning latent trees as part of a deep model through supervision on other tasks, typically via differentiable structured hidden layers \cite{kim2017struct,bradbury2017nmt,liu2018struct,tran2018nmt,peng2018spigot,niculae2018dynamic,liu2018structalign}, 
 policy gradient-based approaches \cite{yogatama2017learning,williams2018latent,havrylov2019}, or differentiable relaxations \cite{choi2018learning,maillard2018latent}.

The variational approximation uses amortized inference \cite{kingma2014vae,mnih2014neural,rezende2014vae}, in which an inference network is used to obtain
the variational posterior for each observed $\boldx$. Since our inference network is structured (i.e., a CRF), it is also related to CRF autoencoders \cite{ammar2014crf} and structured VAEs \cite{johnson2016composing,krishnan2017struct}, which have been used previously for unsupervised \cite{cai2017crf,drozdov2018latent,li2019grammar} and semi-supervised \cite{yin2018structvae,corro2018semi} parsing.
\vspace{-2mm}
\section{Conclusion}
\vspace{-2mm}
It is an open question as to whether explicit modeling of syntax significantly helps  neural models. \citet{strubell2018srl} find that supervising intermediate attention layers with syntactic heads improves semantic role labeling, while \citet{shi2018tree} observe that
for text classification, syntactic trees only have marginal impact. Our work suggests that at least for language modeling, incorporating syntax either via explicit supervision  or as latent variables  does provide useful inductive biases and improves performance. 

Finally, in modeling child language acquisition, the complex interaction of the parser and the grammatical knowledge being acquired is the object of much investigation~\cite{trueswell:2007}; our work shows that apparently grammatical constraints can emerge from the interaction of a constrained parser and a more general grammar learner, which is an intriguing but underexplored hypothesis for explaining human linguistic biases.

{\small
\vspace{-1.5mm}
\section*{Acknowledgments}
\vspace{-1.5mm}
We thank the members of the DeepMind language team for helpful feedback. YK is supported by a Google Fellowship. AR is supported by NSF Career 1845664. }
{
\small
\bibliography{master}
\bibliographystyle{acl_natbib}
}

\end{document}